\documentclass[conference]{IEEEtran}
\IEEEoverridecommandlockouts
\usepackage{arydshln}
\usepackage{cite}
\usepackage{amsmath,amssymb,amsfonts}
\usepackage{algorithmic}
\usepackage{graphicx}
\usepackage{textcomp}
\usepackage{xcolor}
\usepackage{hyperref}
\def\BibTeX{{\rm B\kern-.05em{\sc i\kern-.025em b}\kern-.08em
    T\kern-.1667em\lower.7ex\hbox{E}\kern-.125emX}}
\begin{document}

% \title{Conference Paper Title*\\
% {\footnotesize \textsuperscript{*}Note: Sub-titles are not captured for https://ieeexplore.ieee.org  and
\title{Enhancing Features in Long-tailed Data Using Large Vision Model}
\author{\IEEEauthorblockN{Pengxiao Han}
\IEEEauthorblockA{
\textit{Australian National University}\\
Canberra, Australia \\
pengxiaohan938@gmail.com}
\and
\IEEEauthorblockN{Changkun Ye}
\IEEEauthorblockA{
\textit{Australian National University}\\
Canberra, Australia \\
changkun.ye@anu.edu.au}
\and
\IEEEauthorblockN{Jinguang Tong}
\IEEEauthorblockA{
\textit{Australian National University}\\
Canberra, Australia \\
jinguang.tong@anu.edu.au}
\and
\IEEEauthorblockN{Cuicui Jiang}
\IEEEauthorblockA{
\textit{HKUST}\\
Hong Kong SAR \\
eeccjiang@ust.hk}
\and
\IEEEauthorblockN{\quad\quad Jie Hong\textsuperscript{*}}
\IEEEauthorblockA{
\textit{\quad\quad The University of Hong Kong}\\
        \quad\quad Hong Kong SAR \\
        \quad\quad jiehong@hku.hk}
\and
\IEEEauthorblockN{Li Fang\textsuperscript{*}}
\IEEEauthorblockA{
\textit{Chinese Academy of Sciences}\\
China \\
fangli@fjirsm.ac.cn}
\and
\IEEEauthorblockN{Xuesong Li\textsuperscript{*}\thanks{\textsuperscript{*}Corresponding author}}
\IEEEauthorblockA{
\textit{Australian National University\&CSIRO}\\
Canberra, Australia \\
xuesong.li@csiro.au}
}

\maketitle

\begin{abstract}
Language-based foundation models, such as large language models (LLMs) or large vision-language models (LVLMs), have been widely studied in long-tailed recognition. However, the need for linguistic data is not applicable to all practical tasks. In this study, we aim to explore using large vision models (LVMs) or visual foundation models (VFMs) to enhance long-tailed data features without any language information. Specifically, we extract features from the LVM and fuse them with features in the baseline network's map and latent space to obtain the augmented features. Moreover, we design several prototype-based losses in the latent space to further exploit the potential of the augmented features. In the experimental section, we validate our approach on two benchmark datasets: ImageNet-LT and iNaturalist2018.
\end{abstract}


\section{Introduction}
Long-tailed recognition addresses the class-imbalance problem in practical applications, where dominant head classes have sufficient examples while the tail classes have insufficient ones. This imbalanced distribution is found in many real-world domains, such as species identification~\cite{ miao2021iterative, van2018inaturalist} and medical imaging~\cite{ju2021relational, yang2024spatial}. For example, in species identification tasks, only a few common species are well-represented in datasets, whereas rare species might have minimal instances~\cite{van2018inaturalist}. Similarly, in medical imaging, images of common pathologies are abundant, while images of rare diseases are limited. This imbalance makes it challenging to develop models that perform well across all classes, which results in neural networks that tend to focus on head classes while leading to poor performance on tail classes~\cite{survey1_zhang2024systematic, survey2_zhang2023deep}.

A variety of techniques have been proposed to address the long-tailed recognition problem~\cite{tian2020posterior_labelshift, ye2024labelshift, proco_10444057, li2019detection, ldmlr_han2024latent, vlltr_tian2022vlltrlearningclasswisevisuallinguistic,  BALLAD_ma2021simplelongtailedrecognitionbaseline, cc12m_changpinyo2021conceptual12mpushingwebscale}. Data resampling and augmentation approaches efficiently address the class imbalance by adjusting the frequency of the datasets \cite{ ldmlr_han2024latent, remix_chou2020remix}.
However, the performance of these models relies heavily on the quality of the augmented samples, where the low-quality samples may negatively impact the test accuracy.
Contrastive learning~\cite{contrastive_khosla2020supervised}, which encourages models to learn distinct representations by contrasting pairs, has also been adapted to handle long-tailed data. These contrastive pairs help the model distinguish between head and tail classes more effectively, which enhances feature representations for the tail classes. Some methods design a special neural network to address long-tailed problems~\cite{iscen2021class, li2021self}. However, most of these methods require careful parameter tuning and network design. Feature augmentation is a promising approach to mitigate class imbalance in long-tailed datasets~\cite{hong2022safa, zang2021fasafeatureaugmentationsampling, chu2020featurespaceaugmentationlongtailed}. These methods enable the proposed models to improve additional feature diversity for under-represented classes without requiring explicit additional data or extensive rebalancing techniques. By expanding the feature space of tail classes, feature augmentation techniques help models learn more generalized and discriminative representations. Large language models (LLMs) have demonstrated remarkable generative abilities in recent years. Some works have introduced LLMs to address long-tailed recognition challenges~\cite{cc12m_changpinyo2021conceptual12mpushingwebscale,  BALLAD_ma2021simplelongtailedrecognitionbaseline, xu2024hierarchical}. These works prove their effectiveness in improving model performance on long-tailed distributions.

\begin{figure}
    \centering
    \includegraphics[width=1\linewidth]{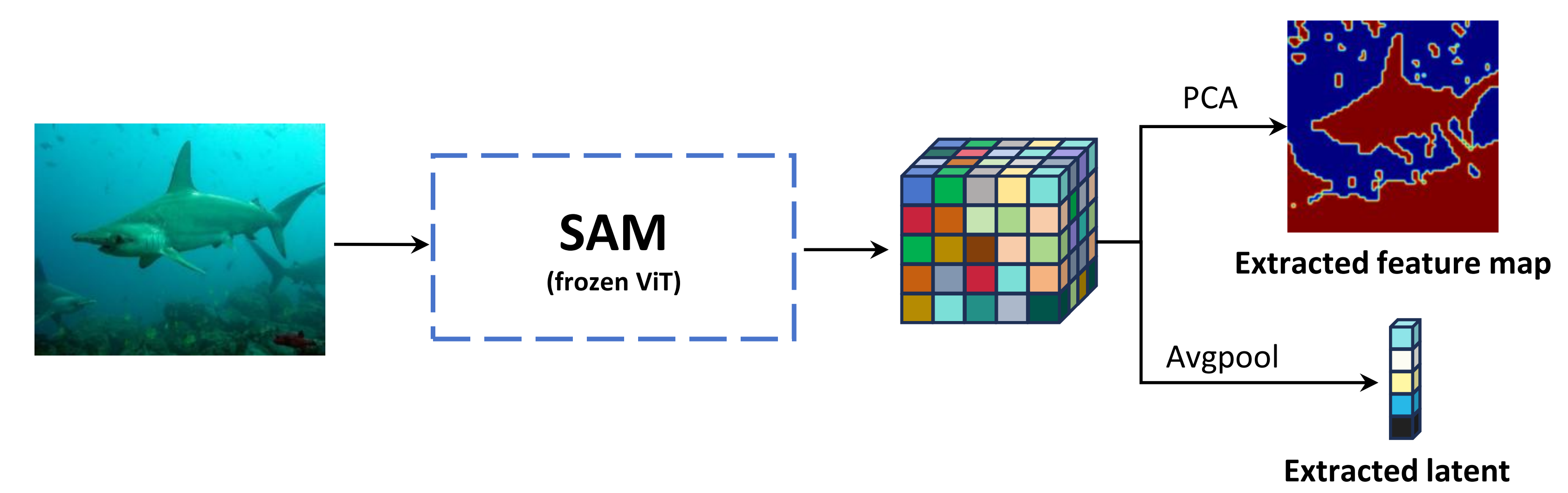}
    \caption{Feature extraction framework of the proposed method. This input image is first processed by the Segment Anything Model (SAM), represented as a frozen Vision Transformer (ViT), to obtain feature output $\mathbf{F}_{SAM}$. Two operations are then applied to utilize the SAM features. In the first approach, Principal Component Analysis (PCA) is used to reduce the input feature channels and output a single-channel feature map. In the second approach, average pooling is applied to generate the SAM feature vector corresponding to the input image.}
    \label{fig:feature_extraction}
\end{figure}

While these previous methods have successfully addressed long-tailed recognition, many rely on external linguistic information to assist feature learning \cite{ma2021simplelongtailedrecognitionbaseline, xu2024hierarchical} or delicate parameter tuning \cite{iscen2021class, hong2022safa}. In this study, we propose a method that leverages large vision models (LVMs) without relying on language-based inputs and uses only visual data to address the long-tailed recognition challenge. Our motivation is to enhance the classification accuracy by maximizing the potential of visual features extracted from LVMs. We introduce a novel feature augmentation strategy that combines LVM-extracted features within the feature map and latent spaces of a baseline network (see Figure~\ref{fig:feature_extraction}). Prototype loss~\cite{protoloss_wen2016discriminative} is a technique in machine learning that focuses on learning a representative prototype for each class in the feature space. This approach is particularly effective for long-tailed recognition problems since prototype loss can help stabilize the learning process by providing a clear target for each class~\cite{wei2022prototype, 10219661, sharma2023learningprototypeclassifierslongtailed}. Prototype loss can enhance the model's generalization ability across both head and tail classes. Therefore, we design a prototype-based loss specifically for long-tailed data, efficiently balancing learning weighting between head and tail classes. By validating our approach on ImageNet-LT~\cite{imagenet18_russakovsky2015imagenetlargescalevisual} and iNaturalist~\cite{van2018inaturalist} datasets, we demonstrate the effectiveness of large vision models in balancing recognition across imbalanced datasets.
The contributions of this work are summarized as follows:
\begin{itemize}
\item We attempt to explore the application of the LVM in long-tailed recognition. The experimental results demonstrate that the use of LVM, to some extent, enhances long-tailed recognition performance. It provides an alternative solution under the long-tailed scenario where the linguistic data is unavailable.
\item We introduce a novel feature augmentation technique that fuses LVM features with baseline network features in both the map and latent spaces, significantly improving imbalanced datasets.
\item Our proposed prototype-based loss dynamically adjusts head and tail classes. The loss ensures compact clustering for head classes while promoting diversity for tail classes, thus improving balance in long-tailed recognition.
\end{itemize}

\section{Related work}
\label{sec:related_work}
\subsection{Long-tailed Recognition}
Long-tailed recognition tackles the challenge of class imbalance, where the majority classes dominate the dataset, and the minority classes are underrepresented. This problem is critical in many computer vision applications, but it poses difficulties because neural networks are prone to bias toward head classes, degrading the performance of tail classes. The primary goal of long-tailed recognition is to improve overall classification performance on imbalanced datasets. In recent years, various methods have been proposed to address this challenge~\cite{survey1_zhang2024systematic, survey2_zhang2023deep}.
Data resampling has proven effective in improving long-tailed recognition performance. LDMLR~\cite{ldmlr_han2024latent} enhances data augmentation by generating pseudo-features in a latent space using a diffusion model~\cite{ddim_song2022denoisingdiffusionimplicitmodels}, which is combined with the original features to fine-tune the classifier.

Contrastive learning~\cite{contrastive_khosla2020supervised,ye2022efficient} has also been explored for long-tailed recognition. ProCo~\cite{proco_10444057} uses contrastive learning and addresses the lack of contrastive pairs by assuming that tail classes follow a von Mises-Fisher (vMF) distribution, which samples contrastive pairs. Another important approach involves architectural design for long-tailed learning. DeiT-LT~\cite{deitlt_rangwani2024deit} adopts a teacher-student framework that distills knowledge from ResNet CNN models~\cite{resnet_he2015deep} into a Vision Transformer (ViT)~\cite{vit_dosovitskiy2021imageworth16x16words} student, allowing the model to better focus on the tail classes. Logit adjustment is another technique used to improve accuracy by rectifying prediction logits according to the frequency of labels in the training data.
MAPLS~\cite{ye2024labelshift} addresses dataset imbalance problems by proposing a Bayesian framework with a Maximum A Posteriori (MAP) model. The method incorporates an EM algorithm for convergence and an Adaptive Prior Learning (APL) model to select prior parameters dynamically. It leverages Markov Chain Monte Carlo (MCMC) sampling for posterior estimation, effectively handling class imbalance without requiring classifier fine-tuning.
\begin{figure*}[t]
    \centering
    \includegraphics[width=1\linewidth]{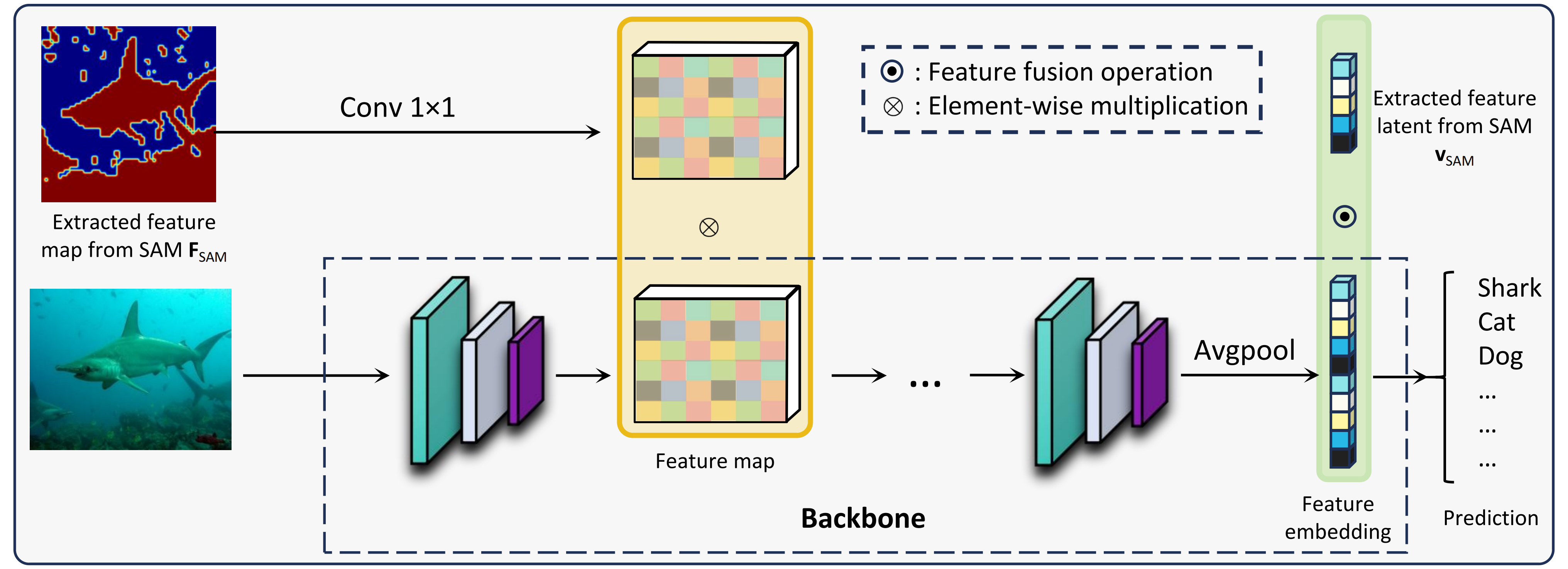}
    \caption{Framework of the proposed method illustrating two feature augmentation methods: feature map fusion and latent feature vector fusion. The extracted feature map from SAM undergoes $1 \times 1$ convolution and is fused with the backbone feature map via element-wise multiplication and addition. Subsequently, the extracted feature latent from SAM is fused with the feature embedding of the backbone after average pooling.}
    \label{fig:framework}
\end{figure*}

\subsection{Foundation Model in Long-tailed Recognition}
Large Language Models (LLMs) have been studied for long-tailed recognition by incorporating external linguistic knowledge to enhance model performance~\cite{ vlltr_tian2022vlltrlearningclasswisevisuallinguistic, BALLAD_ma2021simplelongtailedrecognitionbaseline, cc12m_changpinyo2021conceptual12mpushingwebscale}. Methods using LLMs integrate linguistic information into visual tasks, enabling the models to generalize better to tail classes by understanding class relationships at a conceptual level. VL-LTR~\cite{vlltr_tian2022vlltrlearningclasswisevisuallinguistic} is a novel framework that leverages both visual and linguistic representations to address long-tailed visual recognition. The method consists of two key components: class-wise visual-linguistic pre-training (CVLP) to link image and text representations at the class level and a language-guided recognition (LGR) head to enhance recognition performance by dynamically combining visual and linguistic features. VL-LTR demonstrates high performance, especially in classes with fewer samples. ~\cite{wang2023exploringvisionlanguagemodelsimbalanced} adds a lightweight decoder of vision-language models and incorporates imbalanced learning techniques to improve classification accuracy on long-tailed datasets. BALLAD~\cite{BALLAD_ma2021simplelongtailedrecognitionbaseline} leverages CLIP~\cite{clip_radford2021learningtransferablevisualmodels} to address long-tailed recognition. The method involves two phases: Phase A continues pretraining the vision-language backbone on long-tailed data to update its representations, while Phase B uses a linear adapter to fine-tune the model on balanced data, improving performance on tail classes. LIFT~\cite{shi2024longtaillearningfoundationmodel} proposes a lightweight fine-tuning method with foundation models and achieves faster training and better performance.
Although these methods effectively leverage LLMs to improve long-tailed recognition performance, they often require linguistic information during training and inference, which is unavailable in many situations.

\subsection{Large Vision Model}
LVMs have emerged as an important advancement in computer vision, demonstrating remarkable capabilities across various tasks such as classification, segmentation, and image generation. These models are designed with large-scale architectures and pre-trained on massive datasets, allowing them to generalize effectively to unseen data. One example is the Segment Anything Model (SAM)~\cite{SAM_kirillov2023segment}, which is capable of segmenting objects in images with minimal user interaction and zero-shot generalization to unseen data.

Recent studies have extended SAM's capabilities across different domains. For example, UnSAM~\cite{SAMow_wang2024segmentsupervision} replaces manual annotations with a self-supervised approach to generate hierarchical segmentation masks and achieves competitive results with the supervised counterpart SAM. \cite{dissam_zhang2024distillingsemanticpriorssam} takes SAM's advanced semantic priors for image restoration. The method distills the semantic prior from SAM and uses the prior in lightweight models to improve restoration performance without impacting inference speed.

However, despite SAM's powerful segmentation abilities, its potential to address long-tailed recognition has not been explored. In this work, we leverage SAM's robust generalization capabilities to address the challenges of long-tailed recognition. Our approach begins by applying SAM to segment the training dataset images, followed by extracting features from the segmented images using a convolutional layer. These extracted features are then fused with the original features from a trainable convolutional neural network model. This pipeline effectively utilizes the generalization capabilities of the SAM model, enabling our model to perform well in Long-Tailed recognition problems. By integrating the SAM features into our proposed pipeline, we demonstrate its potential to efficiently and effectively handle the head and the tail classes in the Long-Tailed datasets.

\section{Method}
\label{sec:method}
\begin{figure}[t]
    \centering
    \includegraphics[width=1\linewidth]{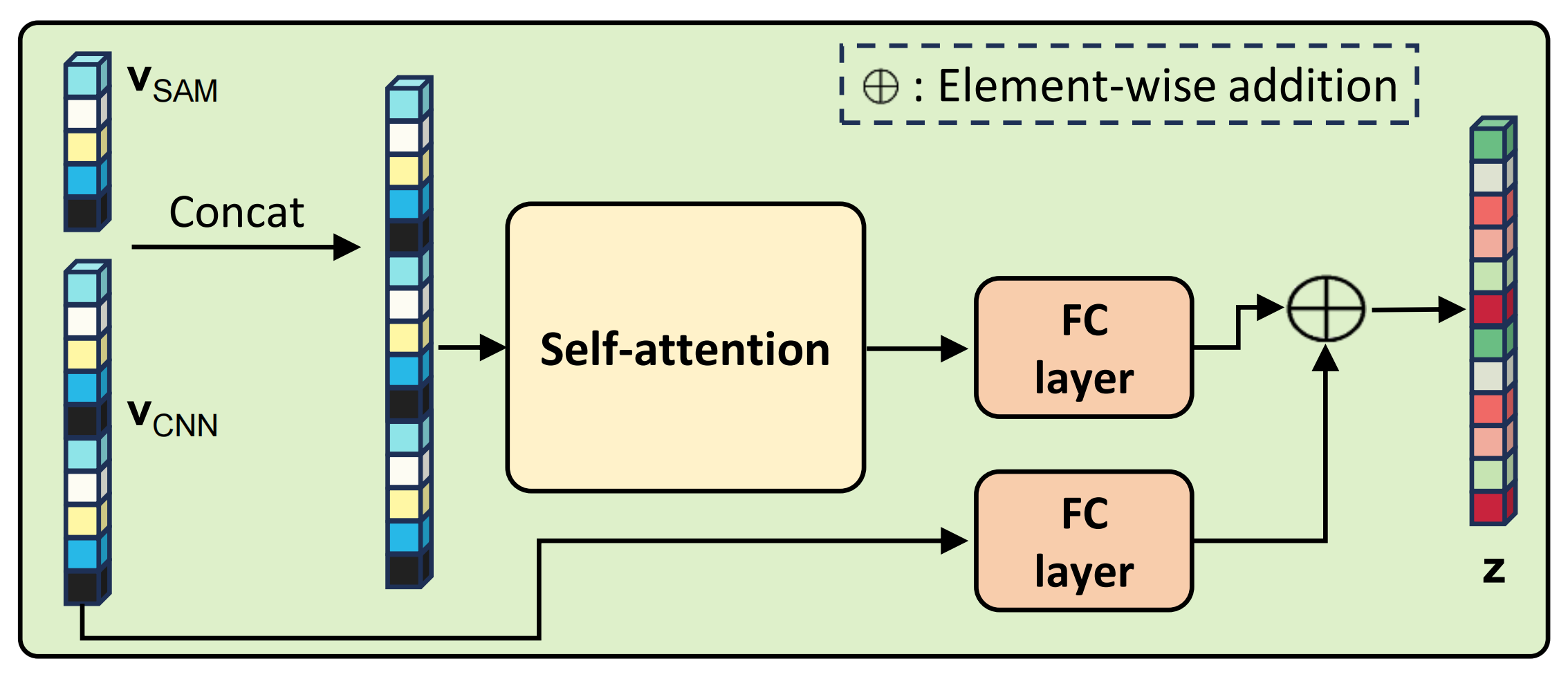}
    \caption{Framework for feature latent fusion into the backbone. The SAM and CNN feature vectors are concatenated and processed through a self-attention module, followed by a fully connected (FC) layer. The resulting features are then fused with the original CNN features through element-wise addition to enhance the final representation.}
    \label{fig:framework_latent}
\end{figure}

\begin{figure}
    \centering
    \includegraphics[width=1\linewidth]{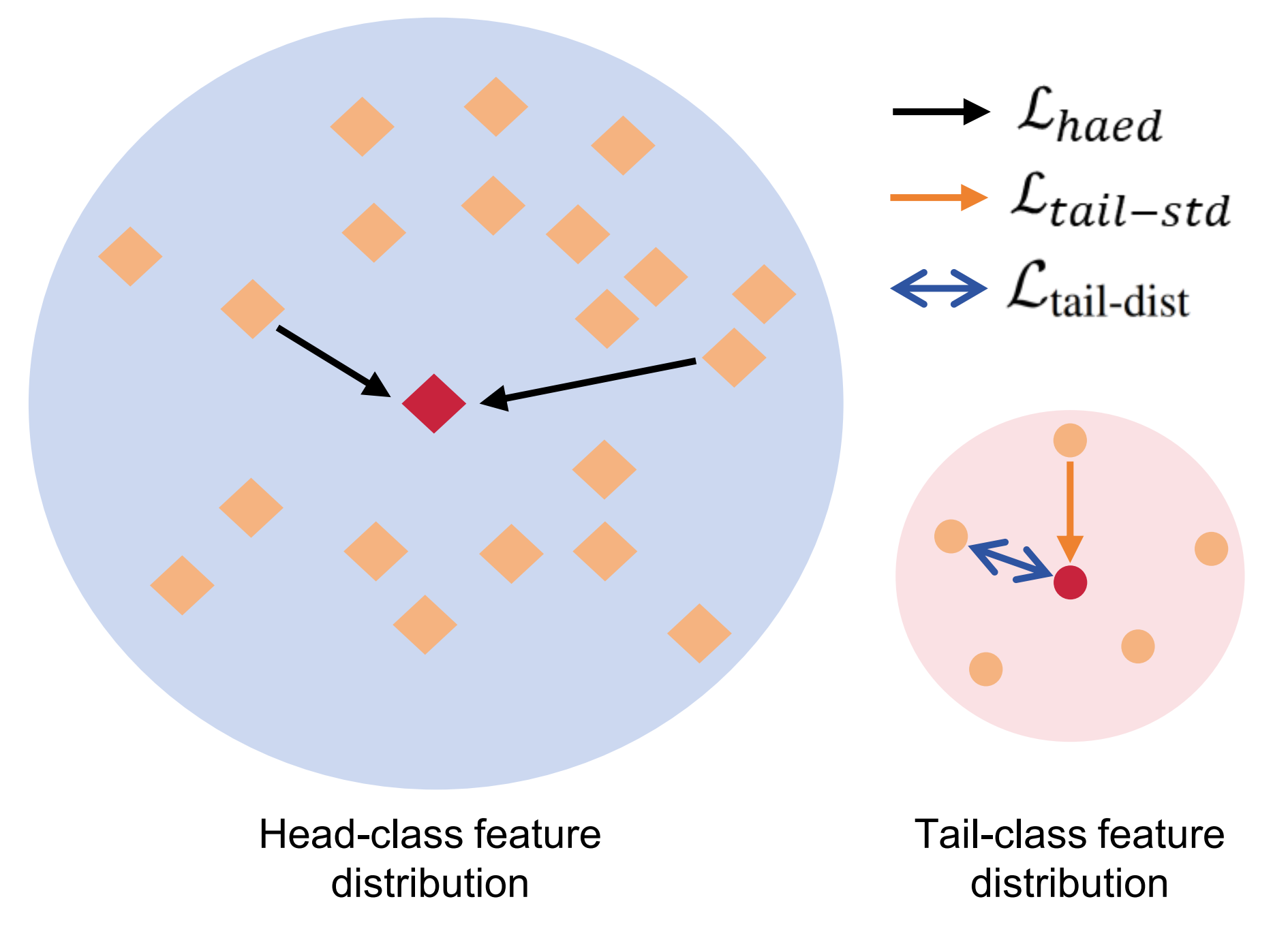}
    \caption{Illustration of the proposed prototype-based loss. The figure consists of three loss components: $\mathcal{L}_{\text{head}}, \mathcal{L}_{\text{tail-std}}$ and $\mathcal{L}_{\text{tail-logdist}}$. The head loss, $\mathcal{L}_{\text{head}}$ (black arrows), encourages comparison clustering around the class prototype for head classes. For tail classes, $\mathcal{L}_{\text{tail-std}}$ (orange arrow) minimizes the distance between features and the prototype, while $\mathcal{L}_{\text{tail-logdist}}$ (blue arrow) enhances diversity by pushing features away from the prototype.}
    \label{fig:loss}
\end{figure}

In this work, our aim is to study the use of LVMs to address the long-tailed recognition problem. In the first part, as shown in Figure~\ref{fig:framework}, we propose to utilize the knowledge extracted from SAM to enhance (or augment) the baseline features. Secondly, the prototype-based loss (see Figure~\ref{fig:loss}) is designed to further exploit the effectiveness of enhanced (or augmented) features by the Segment Anything Model (SAM).
We denote the long-tailed training dataset as \( \mathcal{D} = \{(x_i, y_i)\}_{i=1}^N \), where $x_i$ represents the input image, \( y_i \in \mathcal{Y} \) is the corresponding class label. Let \( \mathcal{F}_{\text{SAM}}(\mathbf{x}) \) denote the feature map extracted from SAM for an input image $\mathbf{x}$.

\subsection{Feature Augmentation Using SAM}\label{subsec:f-aug}
The first step of our method involves extracting features from SAM for the training, validation, and testing datasets. We denote the extracted feature as \( \mathbf{F}_{SAM} \in \mathbb{R}^{C \times H_0 \times W_0} \), where \( C \) is the number of channels, and \( H_0 \) and \( W_0 \) are the height and width of the feature map, respectively. Next, we apply two different operations on the extracted features, corresponding to two different ways of utilizing the SAM features.

\subsubsection{SAM Map Feature}
The SAM features can be used as a mask to help the classifier concentrate on the critical areas in the image.
In this method, we reduce the channels of \( \mathbf{F}_{SAM} \) using Principal Component Analysis (PCA) to create a feature map with a single channel. This transformed feature map is denoted as
$
\mathbf{F}_{\text{PCA}} \in \mathbb{R}^{1 \times H_1 \times W_1},
$
where \( H_1 \) and \( W_1 \) are the new height and width after PCA transformation.

After obtaining the PCA-transformed feature, we apply resizing and normalization operations to reshape it to the dimension of \( 1 \times H_i \times W_i \), denoted as
$
\mathbf{F}_{\text{norm}} \in \mathbb{R}^{1 \times H_i \times W_i}.
$
where $\mathbf{F}_{\text{norm}}=\mathbf{F}_{\text{SAM}}$ is called SAM map feature in our work. Next, we apply a \( 1 \times 1 \) convolution on this feature map, transforming it into a feature map of size \( C_i \times H_i \times W_i \):
\begin{equation}
\mathbf{F}_{\text{conv}} = \text{Conv}_{1 \times 1}(\mathbf{F}_{\text{norm}}) \in \mathbb{R}^{C_i \times H_i \times W_i}.
\end{equation}

This transformed feature is then integrated into the ResNet backbone. Specifically, it is element-wise multiplied with the corresponding feature \( \mathbf{F}_{\text{res}} \) from the ResNet backbone:
\begin{equation}
\mathbf{F}_{\text{fused}} = \mathbf{F}_{\text{conv}} \otimes \mathbf{F}_{\text{res}} \oplus \mathbf{F}_{\text{res}},
\end{equation}
where \( \otimes \) denotes element-wise multiplication. Overall, the SAM feature map enhances the baseline features by providing spatial information.

\subsubsection{SAM Latent Feature}
The SAM features can also be integrated into the image classifier through feature concatenation. In this method, as illustrated in Figure~\ref{fig:framework_latent}, we apply average pooling on the extracted SAM feature to generate a latent vector, denoted as $\mathbf{v}_{\text{SAM}} = \text{AvgPool}(\mathbf{F}_{\text{SAM}})$, where $\mathbf{v}_{\text{SAM}} \in \mathbb{R}^{C}$ is called the SAM latent feature in our work. This latent vector is concatenated with the output of the last convolutional layer of the CNN backbone, which is also of size \( 1 \times 1 \), $\mathbf{v}_{\text{concat}} = [\mathbf{v}_{\text{CNN}}, \mathbf{v}_{\text{SAM}}]$.
Next, the concatenated vector is processed by a self-attention mechanism and a fully connected (FC) layer to produce a logit, which denotes as $\mathbf{z}_{\text{SAM}} = \text{FC}(\text{SelfAttention}(\mathbf{v}_{\text{concat}}))$, where $\mathbf{z}_{\text{SAM}} \in \mathbb{R}^{C}$ is the encoded logit by SAM. The final logits are obtained by element-wise addition with the original CNN logit vector $\mathbf{z}_{\text{CNN}}$, where $\mathbf{z}_{\text{CNN}}$ is the fully connected layer output from $v_{\text{CNN}}$ in Figure ~\ref{fig:framework_latent}:
\begin{equation}
\mathbf{z} = (1-\alpha) \mathbf{z}_{\text{SAM}} + \alpha \mathbf{z}_{\text{CNN}}
\end{equation}
where $\alpha$ is a hyperparameter. This results in the final logit, $\mathbf{z}$, used for classification.

\subsection{Prototype-based Loss}\label{subsec:lt-loss}
As introduced in subsection~\ref{subsec:f-aug}, the augmentation by SAM empowers the baseline feature with extra knowledge. To further exploit the effectiveness of these augmented features, we design the prototype-based loss (see Figure \ref{fig:loss}). The prototype-based loss mainly focuses on exploring the potential of enhanced features from head classes. Besides, the loss is also considered in maintaining the performance of tail classes.

\subsubsection{Center Loss}
Motivated by \cite{protoloss_wen2016discriminative}, we introduce the center loss, which minimizes the distance between feature representations of images and the prototypes representing each class in the latent space. For a given class $k$, the prototype denoted as $\mathbf{c}_k$ represents a centroid vector that mainly captures the feature distribution of class $k$.
Given an input feature matrix \( \mathbf{Z} \in \mathbb{R}^{B \times C} \), where \( B \) is the batch size and \( d \) is the feature dimension, along with ground-truth labels \( \mathbf{y} \in \{1, \ldots, K\}^B \) for \( K \) classes, the center loss is defined as:

\begin{equation}
\mathcal{L}_{\text{center}} = \frac{1}{B} \sum_{i=1}^B \left\| \mathbf{z}_i - \mathbf{c}_{y_i} \right\|^2_2,
\end{equation}
where \( \mathbf{z}_i \) is the feature vector for sample \( i \), and \( \mathbf{c}_{y_i} \) is the prototype feature for the true class \( y_i \).

\subsubsection{Memory-based Prototype}
To compute the prototype $\mathbf{c}_k$, in our work, we assign each class with a memory bank, which includes $M$ elements. The memory bank is dynamically updated during training. After each iteration, the augmented features from the model for the class are appended to the memory bank as new elements.
The earliest stored elements are discarded if the memory bank size exceeds the limit $M$.
This restriction allows the computed prototype to capture information from the most recent features. The prototype of each class can then be obtained by averaging over the elements in the memory bank.

\subsubsection{Prototype-based Head Loss}
The first term of the prototype-based loss is proposed for the head classes. We assume that SAM encodes the rich information within the head-class features. Therefore, the prototype based on the augmented head-class features would become very representative. This loss aims to achieve compact head-class features around their respective class prototypes. The distance between a feature vector $\mathbf{z}_i$ and the class prototype $\mathbf{c}_{y_i}$ can defined as:
\begin{equation}
\text{d}(\mathbf{z}_i, \mathbf{c}_{y_i}) = \|\mathbf{z}_i - \mathbf{c}_{y_i}\|_2^2
\end{equation}

Given a batch of images with a batch size of $B$, the prototype-based head loss is designed as follows:
\begin{equation}
\mathcal{L}_{\text{head}} = \frac{1}{B} \sum_{i=1}^{B} (1 - w_{y_i}) \, \text{d}(\mathbf{z}_i, \mathbf{c}_{y_i}) \mathbb{I}(w_{y_i}<\tau_{\text{head}})
\end{equation}
where $w_{y_i}$ denotes the imbalance weight for class $y_i$. The head classes are assigned with lower weights. $\mathbb{I}(.)$ is the indicator function. Hence, $\mathcal{L}_{\text{head}}$ is only used for head classes whose $w_{y_i}$ are lower than $\tau_{\text{head}}$.

\subsubsection{Prototype-based Tail Loss}
Tail class prototypes cannot be evaluated in the same way as the head class prototypes, since tail classes have few samples in a Long-Tailed training set and the neural network models may not learn reliable feature representations for these classes.
Two terms of the prototype-based loss are proposed to learn the features from tail classes -- the Tail-std loss $\mathcal{L}_{\text{tail-std}}$ and the Tail-dist loss $\mathcal{L}_{\text{tail-dist}}$.

\noindent \textbf{Tail-std Loss.} For the first loss, $\mathcal{L}_{\text{tail-std}}$, to promote feature distribution around the tail-class prototype and mitigate overfitting, we introduce a stochastic perturbation to the class centers:
\begin{align}
\mathbf{c}_{k} &\leftarrow \mathbf{c}_{k} + \sigma_k \text{, or} \\
\mathbf{c}_{k} &\leftarrow \mathbf{c}_{k} - \sigma_k
\end{align}
where $\sigma_k$ is the standard deviation of the elements in the memory bank for class $k$.
The plus or minus of the $\sigma_k$ is chosen at random.
Based on these refined prototypes, we propose the prototype-based tail-std loss:
\begin{equation}
\mathcal{L}_{\text{tail-std}} = \frac{1}{B} \sum_{i=1}^{B} w_{y_i} \, \text{d}(\mathbf{z}_i, \mathbf{c}_{y_i}) \mathbb{I}(w_{y_i}>\tau_{\text{tail-std}})
\end{equation}
where $\mathbb{I}(\cdot)$ is the indicator function. $\tau_{\text{tail-std}}$ is a pre-defined threshold used to identify tail-classes features for computing $\mathcal{L}_{\text{tail-std}}$. $\mathcal{L}_{\text{tail-std}}$ encourages the tail-class features to approximate a distribution around a prototype.

\noindent \textbf{Tail-dist Loss.} The purpose of $\mathcal{L}_{\text{tail-std}}$ is similar to that of head classes: the loss forces the predicted features of the tail classes to concentrate toward a representative prototype, which is the center of the features in the training set. This encourages the model to predict features that are close to this prototype.
To enrich the diversity of the tail-class features, we design the prototype-based tail-dist loss that uses a negative logarithmic distance to push the tail-class features away from the prototype:
\begin{align}
\mathcal{L}_{\text{tail-dist}} =
& -\frac{1}{B} \sum_{i=1}^{B} w_{y_i} \log \left( \text{d}(\mathbf{z}_i, \mathbf{c}_{y_i}) \right) \mathbb{I}(w_{y_i}>\tau_{\text{tail-dist}})
\end{align}
where $\mathbb{I}(.)$ is the indicator function. $\tau_{\text{tail-dist}}$ is a pre-defined threshold used to identify tail-classes features for computing $\mathcal{L}_{\text{tail-dist}}$. The common point among $\mathcal{L}_{\text{tail-std}}$ and $\mathcal{L}_{\text{tail-dist}}$ is we encourage tail-class features to distribute in the larger space.

\noindent \textbf{Total Loss.} The prototype-based loss is the combination of the head and tail-class losses:
\begin{equation}
\mathcal{L}_{\text{proto}} = \mathcal{L}_{\text{head}} +  \mathcal{L}_{\text{tail-std}} + \mathcal{L}_{\text{tail-dist}}
\end{equation}
where $\mathcal{L}_{\text{proto}}$ encourages the model to explore more potential of the enhanced head-class features while maintaining the performance of features from tail classes. It results in better generalization of learned features across the long-tailed data distribution. By the way, the total loss for training the model should be $\mathcal{L}_{baseline}+\beta\mathcal{L}_{\text{proto}}$ where $\beta$ is the coefficient to $\mathcal{L}_{\text{proto}}$.

\subsection{Overall Method}
In general, we proposed a novel Neural Network model that utilizes the SAM model for the Long-Tailed Recognition task. The SAM features are integrated into the image classifier through two proposed feature augmentation methods (Section~\ref{subsec:f-aug}). In the meantime, several novel training losses are also proposed to ensure the head classes and the tail classes can benefit from the SAM features (Section~\ref{subsec:lt-loss}).

\section{Experiments}
\label{sec:exp}
%%%%%%%%%%%%%%%%%%
\begin{table*}[ht]
\centering
\resizebox{0.8\textwidth}{!}{%
\begin{tabular}{l|ccc|c}
\hline
\textbf{Method} & \textbf{Many} & \textbf{Medium} & \textbf{Few} & \textbf{All} \\ \hline

CE & 65.9 & 31.6 & 3.7 & 41.6 \\
\hdashline
+SAM map feature & 66.0 & 35.0 & 6.2 & 43.3 \\
+SAM latent feature & 66.6 & 40.2 & 12.0 & 46.7 \\
+SAM map\&latent feature & 67.1 & 39.9 & 12.2 & 46.8 \\
\textbf{+SAM features+prototype-based loss $\mathcal{L}_{\text{proto}}$ (Ours)} & 67.2 & 40.2 & 12.0 & \textbf{46.9} \\
\hline

Label Shift~\cite{ye2024labelshift} & 60.8 & 46.4 & 18.3 & 48.2 \\
\hdashline
+SAM map feature & 60.3 & 47.5 & 25.8 & 49.6 \\
+SAM latent feature & 61.6 & 49.0 & 32.3 & 51.7 \\
+SAM map\&latent feature & 61.4 & 49.5 & 32.7 & 51.9 \\
\textbf{+SAM features+prototype-based loss $\mathcal{L}_{\text{proto}}$ (Ours)} & 62.7 & 49.5 & 31.8 & \textbf{52.4} \\
\hline

ProCo$^*$~\cite{proco_10444057} &67.7 &54.1 &38.1 &57.2 \\
\textbf{+SAM features+prototype-based loss $\mathcal{L}_{\text{proto}}$ (Ours)}  &\textbf{68.0} &\textbf{56.4} &\textbf{41.1} &\textbf{58.8} \\
\hline
BCL~\cite{zhu2022balanced} &67.6   &54.6   &36.8   &57.2 \\
NCL~\cite{li2022nested}    &68.2   &53.9   &36.3   &57.0 \\
PaCo~\cite{cui2021parametric} &64.4   &55.7   &33.7   &56.0 \\
\hline
\end{tabular}
}
\vspace{6pt}
\caption{Experimental results on ImageNet-LT~\cite{imagenetlt_liu2019large}. ``SAM features'' is equal to ``SAM map\&latent features''. The classification accuracy in \% is provided. The best numbers are bold.} \label{table_imagenet_LT}
\end{table*}

%%%%%%%%%%%%%%%%%
\begin{table*}[h]
\begin{center}
\resizebox{0.8\textwidth}{!}{
\begin{tabular}{l|ccc|c}
\hline
\textbf{Method} &\textbf{Many} &\textbf{Medium} &\textbf{Few} &\textbf{All} \\ \hline
CE                           &76.5 &66.7 &60.0 &65.0  \\
\hdashline
+SAM map feature             &76.6 &65.4 &60.0 &64.3  \\
+SAM latent feature          &76.4 &68.3 &61.7 &66.5  \\
+SAM map\&latent feature     &76.4 &67.9 &62.2 &66.5  \\
\textbf{+SAM features+prototype-based loss $\mathcal{L}_{\text{proto}}$ (Ours)} &\textbf{76.7} &68.7 & 62.3 &\textbf{67.0} \\
\hline

Label Shift~\cite{ye2024labelshift}
                                                  &69.3 &70.3 &70.8 &70.4  \\
\hdashline
+SAM map feature              &68.1 &69.1 &70.1 &69.4  \\
+SAM latent feature           &70.2 &71.4 &71.7 &71.4  \\
+SAM map\&latent feature      &70.7 &71.5 &\textbf{72.6} &71.9  \\
\textbf{+SAM features+prototype-based loss $\mathcal{L}_{\text{proto}}$ (Ours)}   &73.8 &\textbf{72.9} &72.2 &\textbf{72.7}   \\
\hline
BCL~\cite{zhu2022balanced}  &-  &-   &-   &71.8  \\
RIDE~\cite{wang2020long} &70.2  &72.2   &72.7   &72.2  \\
GCL~\cite{li2022long}  &-  &-   &-   &72.0  \\
\hline

\end{tabular}}
\end{center}
\caption{Experimental results on iNaturalist2018~\cite{van2018inaturalist}. ``SAM features'' is equal to ``SAM map\&latent features''. The classification accuracy in \% is provided. The best numbers are bold.} \label{table_iNaturalist}
\end{table*}

%%%%%%%%%%%%%%%%%%%
\begin{table}[htbp]
\begin{center}
\begin{tabular}{l|ccc|c}
\hline
\textbf{Method} &\textbf{Many} &\textbf{Medium} &\textbf{Few} &\textbf{All} \\ \hline
SAM latent feature               &10.07  &0.36   &0.0  &4.10  \\
SAM latent feature+Label Shift   &15.72    &4.86   &0.49 &8.51\\
\hline
CE                           &76.5 &66.7 &60.0 &65.0  \\
+SAM map\&latent feature     &76.4 &67.9 &62.2 &66.5  \\
\hline
Label Shift~\cite{ye2024labelshift}                                               &69.3 &70.3 &70.8 &70.4  \\
+SAM map\&latent feature      &\textbf{70.7} &\textbf{71.5} &\textbf{72.6} &\textbf{71.9}  \\
\hline
\end{tabular}
\end{center}
\caption{Experimental results on using SAM feature only.} \label{sam_feat_only}
\end{table}

\subsection{Implementation}
Our experiments use ResNet-50~\cite{resnet_he2015deep} as the backbone architecture for both the ImageNet-LT~\cite{imagenetlt_liu2019large} and iNaturalist2018~\cite{van2018inaturalist} datasets. We set the dropout rate to $0.5$ to improve the robustness of the model.
The dimension of feature embedding is $2304$.
For ImageNet-LT, we set the weight of prototype-based loss to $1\mathrm{e}{-4}$,
and we use the Stochastic Gradient Descent (SGD) optimizer, with a learning rate of $0.1$ and a momentum of $0.9$.
For iNaturalist2018, the prototype-based loss weight is adjusted to $0.001$, and
the SGD optimizer had a higher learning rate of $0.2$ and a momentum of $0.9$.

\subsection{Datasets}
\noindent \textbf{ImageNet-LT.}
ImageNet-LT dataset~~\cite{imagenetlt_liu2019large} is derived from ImageNet-2012~\cite{imagenet18_russakovsky2015imagenetlargescalevisual}. ImageNet-LT contains $1000$ categories, with a total of $186389$ images. The number of images per class varies significantly, ranging from $5$ to $1280$, creating a challenging long-tailed learning setting.

\noindent \textbf{iNaturalist2018.}
The dataset iNaturalist2018~\cite{van2018inaturalist} presents a real-world, naturally imbalanced distribution of species, offering a large-scale and fine-grained classification task. Having $8142$ species across $437513$ images, iNaturalist 2018 reflects the natural dataset imbalance found in biodiversity data, where some species are rare and others are common.

%%%%%%%%%%%%%%%%%
\begin{table*}[t]
\begin{center}
\resizebox{0.75\linewidth}{!}{%
\begin{tabular}{l|ccc|c}
\hline
\textbf{Method} & \textbf{Many} & \textbf{Medium} & \textbf{Few} & \textbf{All} \\ \hline

CE & 65.9 & 31.6 & 3.7 & 41.6 \\
\hdashline
+SAM features+$\mathcal{L}_{\text{head}}$ &\textbf{67.3} &38.6 &10.8 &46.0 \\
+SAM features+$\mathcal{L}_{\text{tail-std}}$ &67.2 &39.6 &11.6 &46.5 \\
+SAM features+$\mathcal{L}_{\text{tail-dist}}$ &66.9 &39.8 &\textbf{12.0} &46.6 \\
\hdashline

\textbf{+SAM features+prototype-based loss $\mathcal{L}_{\text{proto}}$} &67.2 &\textbf{40.2} &\textbf{12.0} & \textbf{46.9} \\
\hline

Label Shift~\cite{ye2024labelshift} & 60.8 & 46.4 & 18.3 & 48.2 \\
\hdashline
+SAM feature+$\mathcal{L}_{\text{head}}$ & 62.3 & 48.9 & 31.2 & 51.7 \\
+SAM features+$\mathcal{L}_{\text{tail-std}}$ & 61.9 & 48.9 & 32.5 & 51.7 \\
+SAM features+$\mathcal{L}_{\text{tail-dist}}$ &61.5 &48.9 &\textbf{33.2} &51.7 \\
\hdashline
\textbf{+SAM features+prototype-based loss $\mathcal{L}_{\text{proto}}$} &\textbf{62.7} &\textbf{49.5} & 31.8 & \textbf{52.4} \\
\hline
\end{tabular}%
}
\vspace{6pt}
\caption{Ablation study: the prototype-based loss. The effects of using different losses including $\mathcal{L}_{\text{head}}$, $\mathcal{L}_{\text{tail-std}}$, and $\mathcal{L}_{\text{tail-dist}}$ and $\mathcal{L}_{\text{proto}}$, are studied. The results on ImageNet-LT~\cite{imagenetlt_liu2019large} are given. The classification accuracy in \% is provided. The best numbers are bold.} \label{table_imagenet_LT_loss}
\end{center}
\end{table*}

%%%%%%%%%%%%%%%%%
\begin{table*}[t]
\begin{center}
\resizebox{0.75\textwidth}{!}{
\begin{tabular}{l|ccc|c}
\hline
\textbf{Method} &\textbf{Many} &\textbf{Medium} &\textbf{Few} &\textbf{All} \\ \hline

CE                             &76.5 &66.7 &60.0 &65.0  \\
\hdashline
+SAM features+$\mathcal{L}_{\text{head}}$     &\textbf{78.2} &67.5 &58.2 &64.9  \\
+SAM features+$\mathcal{L}_{\text{tail-std}}$   &75.8 &66.9 &61.7 &65.8  \\
+SAM features+$\mathcal{L}_{\text{tail-dist}}$ &75.4 &67.1 &61.1 &65.6  \\
\hdashline
\textbf{+SAM features+prototype-based loss $\mathcal{L}_{\text{proto}}$} &76.7 &\textbf{68.7} &\textbf{62.3} &\textbf{67.0} \\
\hline

Label Shift~\cite{ye2024labelshift} &69.3 &70.3 &70.8 &70.4  \\
\hdashline
+SAM features+$\mathcal{L}_{\text{head}}$ &\textbf{74.4} &\textbf{73.1} &71.4 &72.6  \\
+SAM features+$\mathcal{L}_{\text{tail-std}}$ &69.3 &70.6 &\textbf{72.2} &71.1  \\
+SAM features+$\mathcal{L}_{\text{tail-dist}}$   &69.2 &72.0 &71.6 &71.0 \\
\hdashline
\textbf{+SAM features+prototype-based loss $\mathcal{L}_{\text{proto}}$}   &73.8 &72.9 &\textbf{72.2} &\textbf{72.7}   \\
\hline

\end{tabular}}
\vspace{6pt}
\caption{Ablation study: the prototype-based loss. The effects of using different losses including $\mathcal{L}_{\text{head}}$, $\mathcal{L}_{\text{tail-std}}$, and $\mathcal{L}_{\text{tail-dist}}$ and $\mathcal{L}_{\text{proto}}$, are studied. The results on iNaturalist2018 are given~\cite{van2018inaturalist}. The classification accuracy in \% is provided. The best numbers are bold.}\label{table_iNaturalist_loss}
\end{center}
\end{table*}

\subsection{Results}
\noindent \textbf{ImageNet-LT.}
We evaluate the proposed method on ImageNet-LT~\cite{imagenetlt_liu2019large} dataset, comparing it against several baselines, including Cross-Entropy (CE), Label Shift~\cite{ye2024labelshift} and ProCo~\cite{proco_10444057}.
The results demonstrate the improvements of our feature fusion module and prototype-based loss design.

As shown in Table~\ref{table_imagenet_LT}, incorporating the SAM map and latent features into the baselines consistently improves performance across all metrics. Specifically, fusing the SAM map feature and the baseline CE increases the overall accuracy from $41.6\%$ to $43.3\%$, while fusing the SAM latent feature raises it further to $46.7\%$. Combining both map and latent features extracted from SAM provides a similar improvement to using the latent feature, reaching $46.8\%$. The largest accuracy gain is achieved when applying the prototype-based loss alongside the SAM map and latent features, yielding an overall accuracy of $46.9\%$. The experiment using Label Shift~\cite{ye2024labelshift} as the baseline shows a similar trend. Using the SAM map feature improves the accuracy from $48.2\%$ to $49.6\%$, while using the SAM latent feature boosts it to $51.7\%$. Using the extracted map and latent features from SAM provides a slight additional increase, reaching $51.9\%$ compared to using the SAM latent feature. The best performance using Label Shift is obtained by the designed prototype-based loss, achieving an accuracy of $52.4\%$. Our model also performs well when applied to the ProCo~\cite{proco_10444057} baseline, increasing the overall accuracy from $57.2\%$ to $58.8\%$.

Moreover, our approach shows significant improvements in the few-shot classes. For example, as demonstrated in Figure~\ref{table_imagenet_LT}, the proposed method achieves accuracy in few-shot classes of $31.8\%$, which has the improvement over the baseline Label Shift by $13.5\%$. This phenomenon indicates that the proposed method can effectively balance the imbalanced dataset using the LVM.

\noindent \textbf{iNaturalist2018.}
We further evaluate our approach on iNaturalist2018~\cite{van2018inaturalist}. We examine its effectiveness across the baselines of CE and Label Shift~\cite{ye2024labelshift}. As shown in Table~\ref{table_iNaturalist}, incorporating extracted SAM features and our prototype-based loss leads to improvements. For the baseline CE, fusing the SAM latent feature increases the overall accuracy from $65.0\%$ to $66.5\%$ while utilizing both the SAM map and latent features yields a similar improvement as the SAM latent feature. The best performance using CE is achieved by integrating the SAM feature and the proposed prototype-based loss, which further boosts the accuracy to $67.0\%$. The experiments on the Label Shift baseline show similar results. Adding SAM latent features increases the overall accuracy from $70.4\%$ to $71.4\%$, and the combination of map and latent features from SAM pushes it to $71.9\%$. The highest accuracy for Label Shift is achieved when the prototype-based loss is added, reaching $72.6\%$ and surpassing the baseline by $2.3\%$.

\subsection{Ablation Study}
\noindent \textbf{Fusion Module.}
In this work, we explore two ways of fusing extracted features from SAM with the long-tailed learning baseline model. As reported in Tables~\ref{table_imagenet_LT} and \ref{table_iNaturalist}, the SAM latent feature outperforms the map feature when using the individual one. The results of utilizing the combination of two modules are better than using the single one alone.

\noindent \textbf{SAM Feature Only.}
Our proposed method relies on extracting features from the SAM model, which is trained on massive data. To demonstrate that the performance improvement is primarily due to the effectiveness of our framework rather than reliance on a particular pre-trained vision model, we directly connect the extracted features from SAM to a fully connected layer for classification tasks. As shown in Table~\ref{sam_feat_only}, using the extracted features from SAM directly for classification does not effectively classify images.

\noindent \textbf{Prototype-based Loss.}
To investigate the impact of different components in our prototype-based loss function, we conduct an ablation study on ImageNet-LT, isolating the effects of $\mathcal{L}_{\text{head}}, \mathcal{L}_{\text{tail-std}}$ and $\mathcal{L}_{\text{tail-dist}}$. This study demonstrates how each loss component contributes to overall classification accuracy and handling head, medium, and few-shot classes.

The prototype-based head loss $\mathcal{L}_{\text{head}}$ improves the performance on many classes but negatively impacts the medium and few-shot classes. As shown in Table~\ref{table_imagenet_LT_loss}, the accuracy for many-shot classes increases to $67.3\%$, while the accuracies for medium and few-shot classes are $38.6\%$ and $10.8\%$, respectively. In contrast, the prototype-based tail-std and tail-dist losses, $\mathcal{L}_{\text{tail-std}}$ and $\mathcal{L}_{\text{tail-dist}}$, perform well on medium and few-shot classes but show limited performances on many-shot classes.
Moreover, we observe a 5.3\% improvement in overall accuracy over the CE baseline when combining $\mathcal{L}_{\text{head}}$, $\mathcal{L}_{\text{tail-std}}$, and $\mathcal{L}_{\text{tail-logdist}}$.
For the Label Shift baseline, we see that the $\mathcal{L}_{\text{head}}$ primarily improves the accuracy of many-shot classes but has limited effect on few-shot classes, resulting in an overall accuracy of $51.7\%$. Using $\mathcal{L}_{\text{tail-std}}$ or $\mathcal{L}_{\text{tail-dist}}$ enhances performance on few-shot classes. $\mathcal{L}_{\text{tail-logdist}}$ achieves best results on few-shot classes at $33.2\%$ and maintaining an overall accuracy of $51.7\%$. The best performance is obtained by combining the $\mathcal{L}_{\text{head}}$ and $\mathcal{L}_{\text{tail-std}}$, reaching an overall accuracy of $52.4\%$, a $4.2\%$ improvement over the baseline. This indicates that integrating prototype-based head and tail loss components provides a balanced improvement.

We also study the effects of prototype loss on iNaturalist2018.
Among the prototype losses, using $\mathcal{L}_{\text{head}}$ alone notably improves the many-shot classes but has limited impact on medium and few-shot. Introducing $\mathcal{L}_{\text{tail-std}}$ and $\mathcal{L}_{\text{tail-logdist}}$ enhance the performance for medium and few-shot classes.
Incorporating $\mathcal{L}_{\text{head}}$ and $\mathcal{L}_{\text{tail-logdist}}$ has the best performance of $67.0\%$ and $72.6\%$ using CE and Label Shift, respectively.

\section{Conclusion}
In conclusion, this paper presents an approach to address the long-tailed recognition challenge by leveraging the LVM (\textit{i.e.}, SAM) without relying on language data. Our method fuses SAM-extracted features to enhance the backbone features for better generalization across an imbalanced dataset. Additionally, we introduce a prototype-based loss function to further augment the fused feature. The loss design considers the aspects of both head and tail classes, which balances the learning across imbalanced data distribution.
Experimental results on benchmark datasets, including ImageNet-LT and iNaturalist2018, validate the effectiveness of our approach, showing consistent improvements across all class distributions.

\bibliographystyle{plain}
\bibliography{main}
\end{document}